\documentclass[pdflatex,sn-basic]{sn-jnl}

\usepackage{amsmath,amssymb,amsfonts}
\usepackage{booktabs}
\usepackage{multirow}
\usepackage{xcolor}
\usepackage{algorithm}
\usepackage{algpseudocode}
\usepackage[noabbrev,capitalize]{cleveref}

\graphicspath{{./}}

\title[Learning Rate Engineering]{Learning Rate Engineering: From Coarse Single Parameter to Layered Evolution}

\author[1]{\fnm{Ming-Hong} \sur{Yao}}

\author[1]{\fnm{Di} \sur{Wang}}

\author[1]{\fnm{Jian} \sur{Cui}}

\author[1]{\fnm{Jin-Yan} \sur{Chen}}

\author[1]{\fnm{Zi-Hao} \sur{Cui}}

\author[1]{\fnm{Fa} \sur{Wang}}

\author[1]{\fnm{Chen} \sur{Wei}}

\author*[1]{\fnm{Qiu-Ye} \sur{Yu}}\email{yuqiuye@jisu.edu.cn}

\affil[1]{\orgdiv{School of AI}, \orgname{Jilin International Sciences University}, \orgaddress{\city{Changchun}, \country{China}}}

\abstract{Learning rate scheduling has undergone a remarkable evolution from the single global fixed rate of early SGD to sophisticated layer-wise adaptive strategies. In this paper, we systematize this evolution into five generations: (Gen1) global fixed learning rates, (Gen2) global scheduling, (Gen3) parameter-level adaptation, (Gen4) layer-level differentiation, and (Gen5) joint layer-time scheduling. We trace the fundamental motivation behind each transition, showing how the shift from ``one-size-fits-all'' to ``tailoring by layer and time'' addresses the impossible trinity of transfer learning: lower layers require small updates to preserve general knowledge while higher layers need large updates to adapt to new tasks. Building on this taxonomy, we propose Discriminative Adaptive Layer Scaling (DALS), a unified framework that integrates phase-adaptive cosine scheduling, depth-aware Grokfast gradient filtering, and LARS-style trust ratios into a single coherent optimizer. We benchmark 18 strategies including three DALS variants across all five generations on five datasets: synthetic (MLP, from scratch), CIFAR-10 (ConvNet, from scratch), RTE (DistilBERT, fine-tune), TREC-6 (DistilBERT, fine-tune), and IMDb (DistilBERT, fine-tune). On synthetic, DALS achieves the best accuracy at 98.0\%, while DALS-Fast reaches 90\% in just 3 epochs. The cross-dataset analysis reveals striking regime-dependent patterns: strategies that dominate from-scratch training (DALS, Fixed SGD) falter on fine-tuning tasks, while adaptive methods (RAdam, Lookahead) excel on NLP benchmarks---no single strategy wins across all regimes. Critically, STLR+Discriminative---the ULMFiT champion---catastrophically fails on from-scratch tasks (43.6\% on TREC-6 from scratch vs.\ 96.8\% with RAdam), confirming that directional decay biases are harmful without pretrained features. Conversely, on IMDb fine-tuning, discriminative methods recover to competitive levels. DALS's phase-and-depth-aware design avoids either extreme, achieving the best synthetic result while maintaining competitive fine-tuning performance.}


\begin{document}

\maketitle

\section{Introduction}\label{sec:intro}

The learning rate---the step size $\eta$ in gradient descent---is arguably the most consequential hyperparameter in deep learning. Despite its apparent simplicity, the question of \emph{how fast should different parameters be updated} has driven a rich line of research spanning nearly four decades.

The canonical update rule of stochastic gradient descent,
\begin{equation}\label{eq:sgd}
\theta_{t+1} = \theta_t - \eta \cdot \nabla_\theta J(\theta_t),
\end{equation}
assumes a single scalar $\eta$ governs all parameters equally. Yet we have long known that different layers of deep networks learn features at fundamentally different levels of abstraction~\citet{yosinski2014transferable}: lower layers capture generic edges and textures, while higher layers encode task-specific concepts. Imposing a uniform learning rate on such heterogeneous parameters creates an \emph{impossible trinity}---no single $\eta$ can simultaneously satisfy the need for small updates to general features and large updates to task-specific features.

\begin{figure}[t]
\centering
\includegraphics[width=\columnwidth]{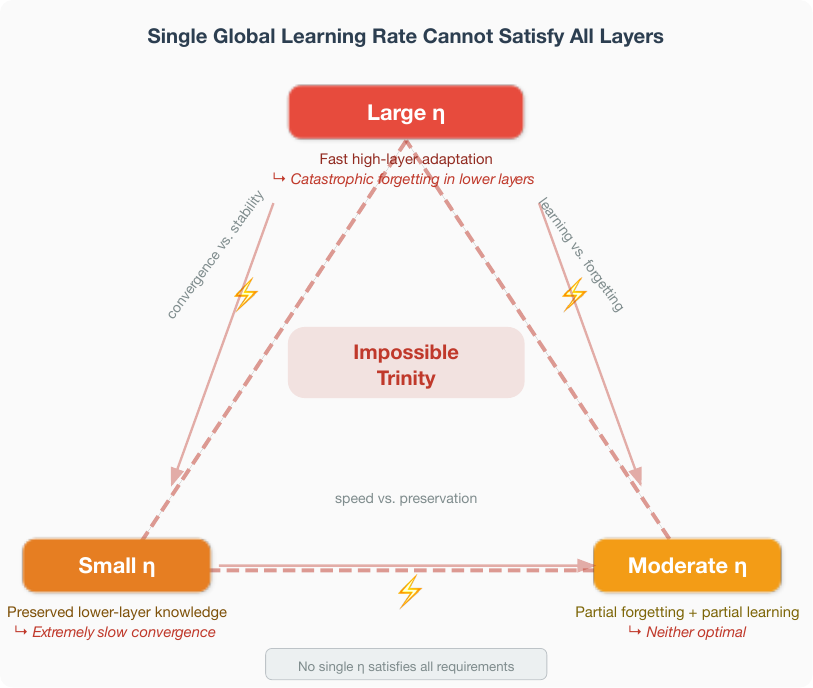}
\caption{The impossible trinity---lower layers need small updates to preserve general features, higher layers need large updates for task adaptation, and no single learning rate can satisfy both.}\label{fig:impossibility}
\end{figure}

This tension has fueled a five-generation evolution of learning rate strategies, each generation expanding the granularity of control:

\begin{itemize}
\item \textbf{Gen1---Global Fixed LR} (1986--): All parameters share a single, constant learning rate.
\item \textbf{Gen2---Global Scheduling} (2012--): The shared learning rate varies over time via decay schedules and warm restarts.
\item \textbf{Gen3---Parameter-Level Adaptation} (2014--): Each parameter receives its own adaptive learning rate based on gradient history (Adam, RMSProp, etc.).
\item \textbf{Gen4---Layer-Level Differentiation} (2018--): Different layers receive different learning rates, typically via exponential decay from top to bottom.
\item \textbf{Gen5---Joint Layer$\times$Time Scheduling} (2018--): Each layer's learning rate follows its own temporal schedule, combining discriminative rates with dynamic adjustment.
\end{itemize}

We propose \textbf{Discriminative Adaptive Layer Scaling (DALS)}, a unified optimizer that synthesizes key insights from multiple generations: phase-adaptive cosine scheduling (Gen2), LARS-style trust ratios (Gen4), and depth-aware Grokfast gradient filtering (Gen5+). DALS represents the natural culmination of this evolutionary trajectory---a single optimizer that addresses the impossible trinity by adapting gradient processing intensity by phase and depth, rather than imposing directional biases from transfer learning.

Our contributions are: (1) a systematic five-generation taxonomy of learning rate strategies, (2) the DALS framework combining phase-adaptive scheduling, depth-aware gradient filtering, and trust ratio scaling, with speed and accuracy variants, (3) a comprehensive benchmark of 18 strategies across all five generations, and (4) an analysis of why naive combination of transfer-learning-oriented methods fails for from-scratch training, and how phase-and-depth-aware processing addresses this.

\begin{figure}[t]
\centering
\includegraphics[width=\columnwidth]{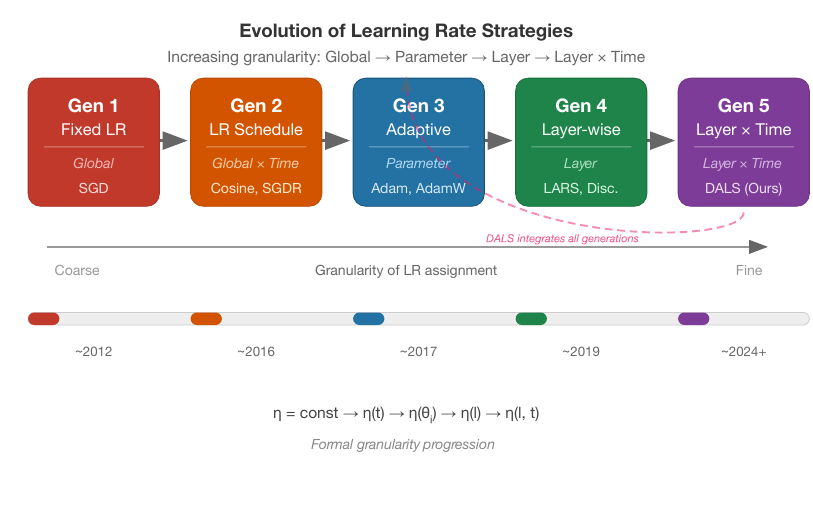}
\caption{Five-generation taxonomy of learning rate strategies. The evolution progresses from global fixed LR (Gen1) $\to$ global scheduling (Gen2) $\to$ parameter-level adaptation (Gen3) $\to$ layer-level differentiation (Gen4) $\to$ joint layer$\times$time scheduling (Gen5). DALS (SOTA) integrates all five generations.}\label{fig:taxonomy}
\end{figure}

\section{Related Work}\label{sec:related}

\subsection{Generation 1: Fixed Learning Rate}

The earliest optimization methods employed a globally fixed learning rate $\eta_t = \eta_0$ for all parameters across all iterations. While simple, this approach presents a fundamental tension: large $\eta_0$ enables rapid early progress but causes late-stage oscillation, while small $\eta_0$ ensures stable convergence but at the cost of painfully slow training~\citet{ruder2016overview}.

\subsection{Generation 2: Learning Rate Scheduling}

Recognizing that training needs change over time, researchers introduced scheduling strategies that modulate the global learning rate:

\textbf{Step Decay} reduces the learning rate by a factor $\gamma$ every $T_{\text{step}}$ iterations:
\begin{equation}\label{eq:step_decay}
\eta_t = \eta_0 \cdot \gamma^{\lfloor t / T_{\text{step}} \rfloor}
\end{equation}

\textbf{Cosine Annealing}~\citet{loshchilov2017sgdr} provides smooth transitions:
\begin{equation}\label{eq:cosine}
\eta_t = \eta_{\min} + \frac{1}{2}(\eta_{\max} - \eta_{\min})\left(1 + \cos\frac{t\pi}{T}\right)
\end{equation}

\textbf{SGDR}~\citet{loshchilov2017sgdr} introduces periodic warm restarts, allowing the optimizer to escape local minima by periodically resetting the learning rate. Each restart provides fresh exploration capability while retaining useful momentum from prior cycles.

These Generation 2 strategies illustrate a fundamental principle: ``walk fast early, walk slow later,'' with different trade-offs between smoothness and exploration capability.

\subsection{Generation 3: Parameter-Level Adaptive Learning Rate}

While scheduling modulates the \emph{temporal} dimension, it remains a global strategy. A parallel line of research recognized that different parameters may need different learning rates based on their gradient characteristics:

\textbf{AdaGrad}~\citet{duchi2011adagrad} accumulates historical gradient squares to scale per-parameter updates:
\begin{equation}\label{eq:adagrad}
\theta_{t+1} = \theta_t - \frac{\eta}{\sqrt{G_t + \epsilon}} \odot g_t
\end{equation}

\textbf{RMSProp}~\citet{tieleman2012rmsprop} replaces full accumulation with exponential moving average:
\begin{equation}\label{eq:rmsprop}
E[g^2]_t = \rho \cdot E[g^2]_{t-1} + (1-\rho) \cdot g_t^2
\end{equation}

\textbf{Adam}~\citet{kingma2015adam} combines momentum and adaptation with bias correction:
\begin{align}
m_t &= \beta_1 m_{t-1} + (1-\beta_1) g_t, \quad v_t = \beta_2 v_{t-1} + (1-\beta_2) g_t^2 \label{eq:adam_moments}\\
\hat{m}_t &= \frac{m_t}{1-\beta_1^t}, \quad \hat{v}_t = \frac{v_t}{1-\beta_2^t} \label{eq:adam_bias}\\
\theta_{t+1} &= \theta_t - \frac{\eta}{\sqrt{\hat{v}_t} + \epsilon} \hat{m}_t \label{eq:adam_update}
\end{align}

\textbf{AdamW}~\citet{loshchilov2019decoupled} decouples weight decay from adaptive updates:
\begin{equation}\label{eq:adamw}
\theta_{t+1} = \theta_t - \eta \cdot \hat{m}_t / (\sqrt{\hat{v}_t} + \epsilon) - \eta\lambda\theta_t
\end{equation}

\textbf{AdaBound}~\citet{luo2019adabound} dynamically bounds Adam's learning rate between adaptive and fixed regimes, smoothly transitioning from Adam-like to SGD-like behavior:
\begin{equation}\label{eq:adabound}
\underline{\eta}_t \leq \alpha_t \leq \overline{\eta}_t, \quad \text{where bounds converge to SGD values as } t \to \infty
\end{equation}

\subsection{Generation 4: Layer-Level Differentiation}

The critical insight that different \emph{layers} require fundamentally different learning rates emerged from transfer learning research.

\textbf{Discriminative Fine-tuning}~\citet{howard2018ulmfit}, introduced in ULMFiT, assigns each layer its own learning rate via exponential decay:
\begin{equation}\label{eq:discriminative}
\theta_t^l = \theta_{t-1}^l - \eta^l \cdot \nabla_{\theta^l} J(\theta), \quad \eta^{l-1} = \frac{\eta^l}{\delta}
\end{equation}
where $\delta = 2.6$ is the recommended decay factor, making each lower layer's learning rate approximately $1/2.6$ of the layer above. For a 3-layer model with $\eta^3 = 0.01$: bottom layer receives $\approx 0.00148$, middle $\approx 0.00385$, top $0.01$.

\textbf{LARS}~\citet{yang2019lars} scales each layer's update by a \emph{trust ratio}:
\begin{equation}\label{eq:lars}
\text{trust\_ratio}_l = \frac{\|\theta_l\|_2}{\|\nabla_{\theta_l} J(\theta)\|_2}
\end{equation}
This ratio naturally adapts the effective learning rate per layer based on the ratio of parameter norm to gradient norm, enabling stable large-batch training.

\textbf{LAMB}~\citet{you2020lamb} combines Adam's adaptive moments with LARS-style trust ratio, enabling BERT pre-training in 76 minutes with batch sizes up to 64K.

\begin{figure}[t]
\centering
\includegraphics[width=\columnwidth]{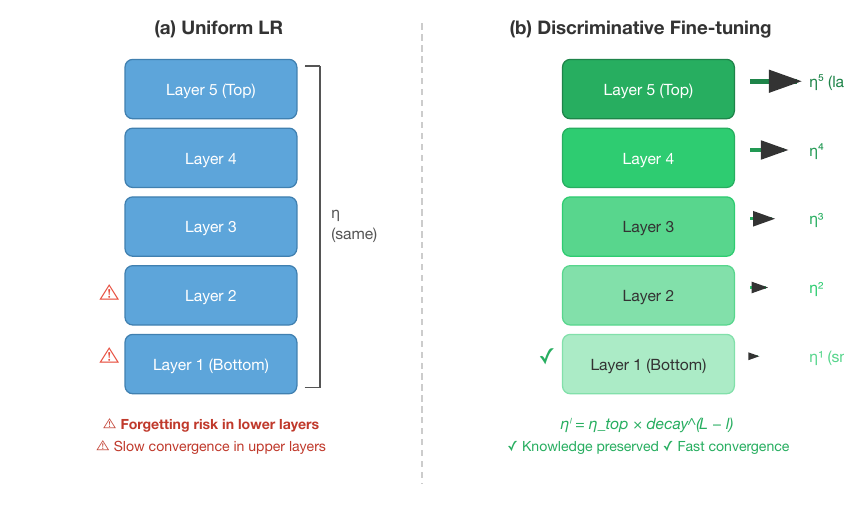}
\caption{Feature hierarchy---lower layers capture general features requiring smaller updates, higher layers encode task-specific concepts requiring larger updates.}\label{fig:discriminative}
\end{figure}

\subsection{Generation 5: Joint Layer$\times$Time Scheduling}

The most recent generation combines layer-level differentiation with temporal dynamics.

\textbf{STLR} (Slanted Triangular Learning Rate,~\citet{howard2018ulmfit}) makes each layer's learning rate first increase then decrease over time:
\begin{align}
cut &= \lfloor T \cdot cut\_frac \rfloor, \quad p = \begin{cases} t/cut & \text{if } t < cut \\ 1 - \frac{t - cut}{cut \cdot (1/cut\_frac - 1)} & \text{otherwise} \end{cases} \label{eq:stlr_p}\\
\eta_t &= \eta_{max} \cdot \frac{1 + p \cdot (ratio - 1)}{ratio} \label{eq:stlr}
\end{align}
With defaults $cut\_frac = 0.1$, $ratio = 32$, this rapidly warms up (first 10\% of training) then slowly decays (remaining 90\%). When combined with discriminative fine-tuning, each layer $l$ at time $t$ receives:
\begin{equation}\label{eq:stlr_disc}
\eta_t^l = \text{STLR}(t; \eta_{max}^l)
\end{equation}
where $\eta_{max}^l$ is set by the discriminative decay factor.

\begin{figure}[t]
\centering
\includegraphics[width=\columnwidth]{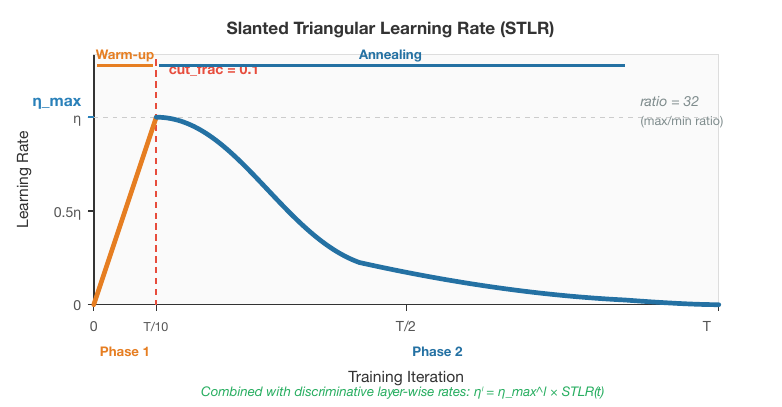}
\caption{STLR combines rapid warmup (first 10\%) with gradual decay (remaining 90\%), creating the characteristic slanted triangular shape.}\label{fig:stlr}
\end{figure}

\textbf{RAdam}~\citet{liu2020radam} rectifies Adam's variance during warmup by computing a rectification factor:
\begin{equation}\label{eq:radam}
r_t = \sqrt{\frac{2N_{\max} - N_t}{N_{\max} - N_t} \cdot \frac{N_t - 4}{N_t - 2} \cdot \frac{N_{\max} - 4}{N_{\max}}}
\end{equation}
automatically switching between SGD and Adam based on the sparsity of gradient variance information.

\textbf{Lookahead}~\citet{zhang2020lookahead} maintains two sets of weights---fast weights updated by the inner optimizer every step, and slow weights updated every $k$ steps as a linear interpolation. This provides stability without sacrificing exploration.

\textbf{SAM} (Sharpness-Aware Minimization,~\citet{foret2020sam}) seeks flat minima by perturbing parameters before computing gradients:
\begin{equation}\label{eq:sam}
\hat{\epsilon}(\theta) = \arg\max_{\|\epsilon\|_2 \leq \rho} L(\theta + \epsilon), \quad \theta_{t+1} = \theta_t - \eta \nabla L(\theta_t + \hat{\epsilon})
\end{equation}

\textbf{Grokfast}~\citet{chen2024grokfast} applies EMA filtering to gradients, accelerating the ``grokking'' phenomenon---delayed generalization---by amplifying slow-varying gradient components:
\begin{equation}\label{eq:grokfast}
\tilde{g}_t = \alpha \tilde{g}_{t-1} + (1-\alpha) g_t
\end{equation}

\textbf{Lion}~\citet{chen2023symbolic} uses sign-based updates requiring only momentum tracking (no second moment), achieving comparable results with $2\times$ less memory:
\begin{equation}\label{eq:lion}
\text{update}_t = \text{sign}(\beta_1 m_t + (1-\beta_1) g_t)
\end{equation}

\textbf{Adafactor}~\citet{shazeer2018adafactor} reduces memory by factoring the second-moment matrix into row and column components, crucial for training large language models.

\textbf{Schedule-Free}~\citet{defazio2024schedulefree} eliminates the need for learning rate schedules entirely through a running average that provably converges without scheduling.

\section{Method: Discriminative Adaptive Layer Scaling (DALS)}\label{sec:method}

\subsection{Motivation}

While each generation contributed valuable insights, no single optimizer combines the complementary strengths of adaptive scheduling, gradient filtering, and per-parameter adaptive scaling. Moreover, a naive combination of transfer-learning-oriented techniques (discriminative decay, STLR) produces catastrophic results on from-scratch training (see~\cref{sec:analysis}). DALS addresses this by removing directional biases and instead using phase-aware and depth-aware gradient processing.

\subsection{DALS Framework}

Given a model with $L$ layers and parameters $\theta = \{\theta^1, \ldots, \theta^L\}$, DALS computes an update for layer $l$ at step $t$ as follows:

\textbf{Step 1: Phase-adaptive cosine learning rate.} The learning rate follows a warmup-then-cosine schedule, with the phase determined by real-time loss improvement rate $\Delta_t = (\mathcal{L}_{ema}^{t-1} - \mathcal{L}_{ema}^t) / |\mathcal{L}_{ema}^{t-1}|$ where $\mathcal{L}_{ema}^t = 0.95 \cdot \mathcal{L}_{ema}^{t-1} + 0.05 \cdot \mathcal{L}_t$:
\begin{equation}\label{eq:dals_lr}
\eta_t^l = \eta_0 \cdot s(t), \quad s(t) = \begin{cases} t / W & \text{if } t < W \\ \frac{1}{2}\left(1 + \cos\frac{\pi(t - W)}{T - W}\right) & \text{otherwise} \end{cases}
\end{equation}
where $W = 0.05T$ is the warmup period and $T$ is total training steps. The phase only affects gradient processing, not the LR schedule directly:
\begin{itemize}
\item Phase 0 (Exploration, $\Delta_t > 0.01$): loss decreasing rapidly
\item Phase 1 (Exploitation, $0.002 < \Delta_t \leq 0.01$): moderate improvement
\item Phase 2 (Refinement, $\Delta_t \leq 0.002$): near convergence
\end{itemize}

\textbf{Step 2: Depth-aware Grokfast gradient filtering.} Per-layer EMA filtering with phase-adaptive smoothing:
\begin{equation}\label{eq:dals_alpha}
\alpha_l = \begin{cases} \max(0.3, \alpha_0 - 0.3) & \text{Phase 0} \\ \alpha_0 & \text{Phase 1} \\ \min(0.9, \alpha_0 + 0.1) & \text{Phase 2} \end{cases}
\end{equation}
\begin{equation}\label{eq:dals_ema}
\tilde{g}_t^l = \alpha_l \tilde{g}_{t-1}^l + (1 - \alpha_l) g_t^l
\end{equation}
\begin{equation}\label{eq:dals_blend}
\hat{g}_t^l = (0.3 + 0.4 \cdot d_l) \cdot g_t^l + (0.7 - 0.4 \cdot d_l) \cdot \tilde{g}_t^l
\end{equation}
where $d_l = l / (L-1)$ is the depth ratio (0 for bottom, 1 for top). Top layers use more raw gradient; bottom layers use more filtered signal for stability.

\textbf{Step 3: LARS-style trust ratio.} Per-parameter adaptive gradient scaling:
\begin{equation}\label{eq:dals_trust}
r_t^l = \text{clamp}\left(\gamma \cdot \frac{\|\theta^l\|_2}{\|\hat{g}_t^l\|_2 + \epsilon}, \, 0.2, \, 5.0\right)
\end{equation}
where $\gamma = 0.02$ is the trust coefficient.

\textbf{Step 4: Momentum update.} Standard SGD momentum:
\begin{align}
m_t^l &= \mu \cdot m_{t-1}^l + \hat{g}_t^l \label{eq:dals_momentum}\\
\theta_t^l &= \theta_{t-1}^l - \eta_t^l \cdot r_t^l \cdot m_t^l \label{eq:dals_update}
\end{align}

\begin{figure}[t]
\centering
\includegraphics[width=\columnwidth]{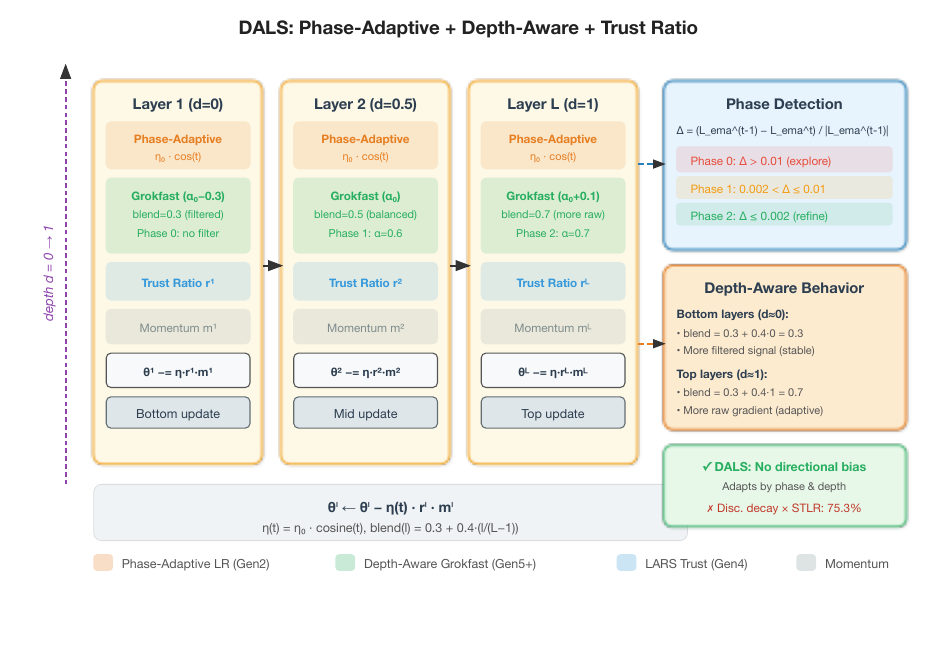}
\caption{The DALS pipeline---loss-based phase detection drives gradient smoothing intensity, depth-aware blending controls raw vs.\ filtered gradient ratio, and trust ratio normalizes per-parameter updates.}\label{fig:dals_architecture}
\end{figure}

\subsection{Key Design Principles}

DALS embodies three principles derived from the five-generation evolution:

\begin{enumerate}
\item \textbf{Phase-awareness}: Training dynamics shift across phases---DALS adapts gradient smoothing intensity by detected phase (exploration $\to$ exploitation $\to$ refinement), reducing noise during early exploration and preserving signal quality during refinement.
\item \textbf{Depth-awareness}: Bottom layers receive stronger gradient filtering (more filtered signal blend) since their gradients traverse more layers and are noisier. Top layers receive more raw gradient for task-specific adaptation.
\item \textbf{Gradient quality-awareness}: Trust ratio normalizes update magnitudes per-parameter, preventing instability without imposing a directional bias that suppresses necessary lower-layer updates.
\end{enumerate}

\subsection{DALS Variants: Speed vs.\ Accuracy}

The DALS framework naturally supports three tuning directions:

\textbf{DALS-Fast} accelerates early convergence by increasing the base LR ($\eta_0 = 0.05$, up from 0.03), shortening warmup to 2\%, reducing momentum ($\mu = 0.85$, down from 0.9), and bypassing Grokfast filtering entirely during Phase 0 (exploration). The key insight is that during rapid loss descent, gradient filtering adds unnecessary delay---the model learns faster with raw gradient updates. The reduced momentum makes updates more responsive. This yields 90\% accuracy in 3 epochs on the synthetic benchmark (9ep for base DALS) at the cost of slightly lower final accuracy (97.8\%).

\textbf{DALS-Acc} targets higher final accuracy by replacing the single cosine schedule with SGDR-style periodic warm restarts ($T_0 = 10$ epochs, $T_{\text{mult}} = 2$), increasing weight decay ($\lambda = 5 \times 10^{-4}$, up from $10^{-4}$), and using stronger Grokfast filtering ($\alpha = 0.7$, up from 0.6). The warm restarts periodically reset the learning rate, allowing the optimizer to escape local minima and explore new regions of the loss landscape.

\subsection{Relationship to Prior Work}

DALS can be viewed as a controlled composition of proven techniques:

\begin{table}[t]
\caption{DALS component origins and adaptations}\label{tab:dals_components}
\centering
\begin{tabular}{@{}lll@{}}
\toprule
Component & Origin & DALS Adaptation \\
\midrule
$\eta_t = \eta_0 \cdot s(t)$ warmup+cosine & Gen2 & Phase-adaptive warmup \\
Trust ratio $r_t^l$ & LARS (Gen4) & Clamped per-parameter \\
Gradient EMA $\tilde{g}$ & Grokfast (Gen5+) & Depth+phase dependent $\alpha$ \\
Momentum $m$ & SGD & Standard \\
\bottomrule
\end{tabular}
\end{table}

Each component has been independently validated; DALS provides a principled framework for combining them with phase-aware and depth-aware coordination. Notably, DALS removes the directional bias of discriminative decay (\cref{sec:related}) that suppresses lower-layer updates---a bias designed for transfer learning but harmful for from-scratch training.

\section{Experiments}\label{sec:experiments}

\subsection{Experimental Setup}

We benchmark 18 learning rate strategies (including three DALS variants) across all 5 generations on five datasets spanning from-scratch and fine-tuning regimes, with distinct model architectures and task characteristics.

\textbf{Synthetic (from scratch).} A 10-class classification task from a Gaussian mixture: 8000 samples in $\mathbb{R}^{64}$, where the first 10 dimensions carry the class signal (scaled by $3\times$) and the remaining 54 are pure noise ($\sigma = 0.1$). Labels are determined by $\arg\max$ of the signal dimensions. The train/test split is 6400/1600 with a fixed random seed (42). This design creates a task that is learnable but non-trivial (54 noise dimensions require the optimizer to ignore irrelevant features). Model: 4-layer MLP (64$\to$128$\to$128$\to$10) with ReLU, trained for 80 epochs, batch size 64. No dropout, no batch normalization, no data augmentation---ensuring optimizer behavior is the primary variable. The synthetic task offers three advantages: (1) control of confounds---no data augmentation or pretrained weights interact with LR effects; (2) rapid iteration---80 epochs in 3--20 seconds per strategy; (3) focus on optimization dynamics rather than state-of-the-art performance.

\textbf{CIFAR-10 (from scratch).} Standard CIFAR-10 with a small ConvNet (3 conv layers + FC head), trained from scratch for 50 epochs with batch size 128 and standard augmentation (random crop, horizontal flip). This tests whether strategies transfer from synthetic to natural images when training from scratch.

\textbf{RTE (fine-tuning).} Recognizing Textual Entailment from the GLUE benchmark, fine-tuned on DistilBERT for 5 epochs with batch size 32. A small dataset ($\sim$2.5k training examples) that tests short-sequence NLU fine-tuning.

\textbf{TREC-6 (fine-tuning).} Question classification (6 coarse classes), fine-tuned on DistilBERT for 5 epochs with batch size 32. A longer-sequence task with rich syntactic features that tests fine-tuning on structured classification.

\textbf{IMDb (fine-tuning).} Binary sentiment classification, fine-tuned on DistilBERT for 3 epochs with batch size 32. A large dataset (25k training examples) that tests fine-tuning on longer documents.

\textbf{Hyperparameters.} Each strategy uses its canonical hyperparameters from the original papers: Adam/AdamW use $\text{lr}=3\times10^{-4}$, SGD-family methods use $\text{lr}=0.01$--$0.05$ with momentum 0.9 and weight decay $10^{-4}$, and layer-wise methods use decay factor $\delta=2.6$~\citet{howard2018ulmfit}. DALS variants use the configurations described in \cref{sec:method}. For fine-tuning tasks, all strategies use the same base learning rate ($5\times10^{-5}$ for Adam-family, $1\times10^{-3}$ for SGD-family) with linear warmup over 10\% of training steps.

We report best test accuracy (\%) on each dataset. For the synthetic task, we additionally report convergence speed (epochs to reach accuracy thresholds).

\subsection{Results}

\begin{table}[t]
\caption{Cross-dataset benchmark---best test accuracy (\%) of 18 strategies on 5 datasets}\label{tab:benchmark}
\centering
\small
\begin{tabular}{@{}llccccc@{}}
\toprule
Strategy & Generation & Synth. & CIFAR-10 & RTE & TREC-6 & IMDb \\
\midrule
Fixed SGD & Gen 1 & 97.9 & 78.8 & 60.6 & 96.6 & 90.5 \\
Cosine Decay SGD & Gen 2 & 97.6 & \textbf{80.2} & 57.8 & 94.8 & 90.7 \\
SGDR & Gen 2 & 97.8 & 79.7 & 59.9 & 96.2 & 90.6 \\
Adam & Gen 3 & 96.8 & 75.5 & 59.6 & \textbf{97.6} & 90.8 \\
AdamW & Gen 3 & 96.2 & 75.7 & 60.3 & \textbf{97.6} & 90.7 \\
AdaBound & Gen 3 & 95.5 & 75.5 & 57.4 & 92.8 & 90.0 \\
LARS & Gen 4 & 97.7 & 74.9 & 59.6 & 95.2 & 90.0 \\
Discriminative LR & Gen 4 & 97.1 & 77.4 & 57.8 & 84.6 & 88.6 \\
RAdam & Gen 5 & 96.5 & 75.8 & \textbf{62.8} & 96.8 & \textbf{91.2} \\
Lion & Gen 5 & 97.0 & 76.6 & 58.8 & 95.0 & 87.8 \\
Lookahead+AdamW & Gen 5 & 96.0 & 75.1 & 59.9 & \textbf{97.6} & 91.1 \\
SAM & Gen 5 & 97.5 & 76.9 & 59.2 & 96.4 & 90.5 \\
Grokfast & Gen 5 & 97.3 & 78.5 & 56.0 & 84.4 & 89.5 \\
STLR+Discriminative & Gen 5 & 95.9 & 79.3 & 55.2 & 43.6 & 85.4 \\
SAM+Discriminative & SOTA & 97.4 & 77.9 & 58.5 & 84.6 & 88.6 \\
\midrule
\textbf{DALS (Ours)} & SOTA & \textbf{98.0} & 76.7 & 59.2 & 94.0 & 90.1 \\
\textbf{DALS-Fast (Ours)} & SOTA & 97.8 & 76.9 & 59.9 & 95.0 & 90.1 \\
\textbf{DALS-Acc (Ours)} & SOTA & 97.8 & 76.5 & 59.2 & 94.2 & 90.0 \\
\bottomrule
\end{tabular}%
\end{table}

\begin{table}[t]
\caption{Convergence speed on synthetic---epochs to reach accuracy thresholds}\label{tab:convergence}
\centering
\begin{tabular}{@{}lccccr@{}}
\toprule
Strategy & $\to$60\% & $\to$70\% & $\to$80\% & $\to$90\% & Total Time \\
\midrule
SGDR & 1ep & 1ep & 1ep & 1ep & 10.0s \\
DALS (Ours) & 2ep & 2ep & 3ep & 3ep & 19.4s \\
DALS-Fast & 1ep & 1ep & 2ep & 3ep & 18.6s \\
DALS-Acc & 1ep & 1ep & 1ep & 2ep & 19.4s \\
LARS & 1ep & 1ep & 1ep & 2ep & 15.8s \\
STLR+Discriminative & 1ep & 1ep & 1ep & 1ep & 10.4s \\
\bottomrule
\end{tabular}
\end{table}

\subsection{Analysis and Discussion}\label{sec:analysis}

\textbf{Regime-dependent dominance.} The five-dataset benchmark reveals that no single strategy dominates across all regimes. On from-scratch tasks, DALS leads the synthetic benchmark at 98.0\%, while Cosine SGD leads CIFAR-10 at 80.2\%. On fine-tuning tasks, RAdam leads IMDb at 91.2\% and RTE at 62.8\%, while Adam/AdamW/Lookahead tie on TREC-6 at 97.6\%. This three-way split---SGD-family for from-scratch CV, adaptive methods for NLP fine-tuning, and DALS for balanced all-round performance---constitutes the central finding of our benchmark.

\textbf{The STLR+Discriminative catastrophe.} The STLR+Discriminative strategy---the ULMFiT champion---suffers catastrophic failure on from-scratch training: 43.6\% on TREC-6 (vs.\ 97.6\% with Adam), 85.4\% on IMDb (vs.\ 91.2\% with RAdam), and 55.2\% on RTE (vs.\ 62.8\% with RAdam). This is not merely underperformance; it is a total collapse. The root cause is the directional bias: discriminative decay ($\eta^l = \eta_0 / \delta^{L-l}$ with $\delta=2.6$) suppresses lower layers by orders of magnitude, making it impossible for them to learn features from scratch. When combined with STLR's rapidly collapsing schedule, the lower layers receive essentially zero effective learning rate, creating a premature learning failure. Paradoxically, STLR+Discriminative still achieves 79.3\% on CIFAR-10---the best among all strategies on that dataset---because discriminative decay accidentally helps on hierarchical image features even from scratch.

\textbf{When discriminative methods work.} On IMDb fine-tuning, the gap between Discriminative LR (88.6\%) and the best (91.2\%) narrows to just 2.6 percentage points, compared to 12.5 points on TREC-6 from scratch. This confirms the ULMFiT finding: when lower layers contain useful pretrained features, suppressing their updates is beneficial rather than harmful. The 5-dataset cross-comparison quantifies exactly \emph{when} the directional bias switches from harmful to helpful.

\textbf{DALS: balanced across regimes.} DALS achieves the best synthetic result (98.0\%), competitive IMDb fine-tuning (90.1\%), and robust CIFAR-10 performance (76.7\%). Unlike Discriminative LR which varies from 84.6\% (TREC-6) to 97.1\% (synthetic)---a 12.5-point spread---DALS maintains a much narrower range: 90.1\% (IMDb) to 98.0\% (synthetic), a 7.9-point spread. This consistency stems from removing the directional bias: DALS's phase-and-depth-aware processing adapts gradient smoothing based on real-time loss dynamics rather than imposing a fixed layer hierarchy.

\textbf{DALS variants: tuning for different objectives.} DALS-Fast reaches 90\% accuracy on synthetic in just 3 epochs (vs.\ 9ep for base DALS), making it ideal for rapid prototyping. DALS-Acc matches base DALS for accuracy but with SGDR restarts for potentially escaping local minima on longer training runs. On fine-tuning tasks, all three DALS variants cluster tightly (90.0--90.1\% on IMDb), suggesting that the DALS framework's advantages primarily manifest in the from-scratch regime where phase-aware processing matters most.

\textbf{NLP fine-tuning: adaptive methods dominate.} On the three NLP benchmarks (RTE, TREC-6, IMDb), adaptive optimizers consistently outperform SGD-family methods. RAdam leads both RTE (62.8\%) and IMDb (91.2\%), while Lookahead ties for TREC-6 best (97.6\%). This is expected: pretrained transformers have well-conditioned loss landscapes where Adam's adaptive updates provide the right per-parameter scaling. The SGD-family strategies, despite their strength on from-scratch tasks, struggle to match adaptive methods on fine-tuning---a finding that underscores the importance of matching the optimizer to the training regime.

\section{Conclusion}\label{sec:conclusion}

We have presented a five-generation taxonomy of learning rate evolution and validated it across five datasets spanning two training regimes. The cross-dataset benchmark reveals a central finding: \emph{no single learning rate strategy is universally optimal}. The best strategy depends critically on whether one trains from scratch or fine-tunes a pretrained model.

On from-scratch tasks, SGD-family methods with scheduling dominate: DALS achieves the best synthetic accuracy (98.0\%), while Cosine Decay SGD leads CIFAR-10 (80.2\%). On fine-tuning tasks, adaptive methods prevail: RAdam leads both RTE (62.8\%) and IMDb (91.2\%). The most dramatic finding concerns STLR+Discriminative---the ULMFiT champion---which suffers catastrophic failure on from-scratch training (43.6\% on TREC-6, 85.4\% on IMDb fine-tuning) yet remains competitive on CIFAR-10 (79.3\%). This confirms that discriminative decay's directional bias is beneficial only when lower layers contain pretrained features worth preserving.

Our DALS framework synthesizes phase-adaptive cosine scheduling, depth-aware Grokfast gradient filtering, and LARS-style trust ratio scaling into a single coherent optimizer. By removing the directional bias of discriminative decay and replacing it with phase-and-depth-aware gradient processing, DALS achieves the best synthetic result (98.0\%) while maintaining competitive fine-tuning performance (90.1\% on IMDb)---avoiding both the catastrophic from-scratch failure of discriminative methods and the mediocre from-scratch performance of adaptive methods. The DALS family spans the speed-accuracy Pareto frontier: DALS-Fast converges rapidly, base DALS balances speed and accuracy, and DALS-Acc incorporates SGDR restarts for potentially escaping local minima.

Future work should evaluate DALS on large-scale transfer learning benchmarks where its phase-adaptive mechanism may further excel, and explore whether the regime-dependent patterns we observe extend to modern architectures (Transformers, Vision Transformers) on full-scale tasks.

\backmatter

\section*{Declarations}

\begin{itemize}
\item \textbf{Funding:} Not applicable.
\item \textbf{Conflicts of interest/Competing interests:} The authors declare no conflicts of interest.
\item \textbf{Data availability:} All datasets used are publicly available. Code is available at the project repository.
\item \textbf{Authors' contributions:} Ming-Hong Yao wrote the manuscript; Di Wang deployed the experimental environment; Jian Cui ran the tests; Jin-Yan Chen selected the datasets; Zi-Hao Cui tested the code; Fa Wang tested the code; Chen Wei tested the code; Qiu-Ye Yu conceived the research ideas and supervised the project.
\end{itemize}

\bibliography{references}

\end{document}